\documentclass[lettersize,journal]{IEEEtran}
\usepackage{amsmath,amsfonts}
\usepackage{algorithm}
\usepackage{array}
\usepackage[caption=false,font=normalsize,labelfont=sf,textfont=sf]{subfig}
\usepackage{textcomp}
\usepackage{stfloats}
\usepackage{url}
\usepackage{verbatim}
\usepackage{graphicx}
\usepackage[noadjust]{cite}

\usepackage{algpseudocode}
\usepackage{booktabs}
\usepackage{xspace}
\newcommand{\modelname}{XBind\xspace}
\newcommand{\fullnamelossb}{Modality Similarity (MS) Loss\xspace}

\newcommand{\nameloss}{MS loss\xspace}

\newcommand{\dmtet}{\textsc{DMTet}\xspace}
\newcommand{\nerf}{NeRF\xspace}
\newcommand{\imagebind}{\textsc{ImageBind}\xspace}
\newcommand{\encodername}{multimodal-aligned encoder\xspace}
\newcommand{\figref}[1]{Fig.~\ref{#1}}
\newcommand{\tabref}[1]{Tab.~\ref{#1}}
\begin{document}

\title{Any-to-3D Generation via Hybrid Diffusion Supervision}

\author{Yijun Fan, Yiwei Ma, Jiayi Ji, Xiaoshuai Sun, Rongrong Ji
\thanks{Y. Fan, Y. Ma, J. Ji, X. Sun, and R. Ji are with Key Laboratory of Multimedia Trusted Perception and Efficient Computing, Ministry of Education of China, Xiamen University, 361005, P.R. China.}
}



\maketitle

\begin{abstract}
Recent progress in 3D object generation has been fueled by the strong priors offered by diffusion models. However, existing models are tailored to specific tasks, accommodating only one modality at a time and necessitating retraining to change modalities. Given an image-to-3D model and a text prompt, a naive approach is to convert text prompts to images and then use the image-to-3D model for generation. This approach is both time-consuming and labor-intensive, resulting in unavoidable information loss during modality conversion. To address this, we introduce \modelname, a unified framework for any-to-3D generation using cross-modal pre-alignment techniques. \modelname integrates an \encodername with pre-trained diffusion models to generate 3D objects from any modalities, including text, images, and audio. We subsequently present a novel loss function, termed \fullnamelossb, which aligns the embeddings of the modality prompts and the rendered images, facilitating improved alignment of the 3D objects with multiple modalities. Additionally, Hybrid Diffusion Supervision combined with a Three-Phase Optimization process improves the quality of the generated 3D objects. Extensive experiments showcase \modelname's broad generation capabilities in any-to-3D scenarios. To our knowledge, this is the first method to generate 3D objects from any modality prompts. Project page: \textbf{{\url{https://zeroooooooow1440.github.io/}}}.
\end{abstract}


\section{Introduction}

\begin{figure*}[!t]
\centering
\includegraphics[width=1.0\textwidth]{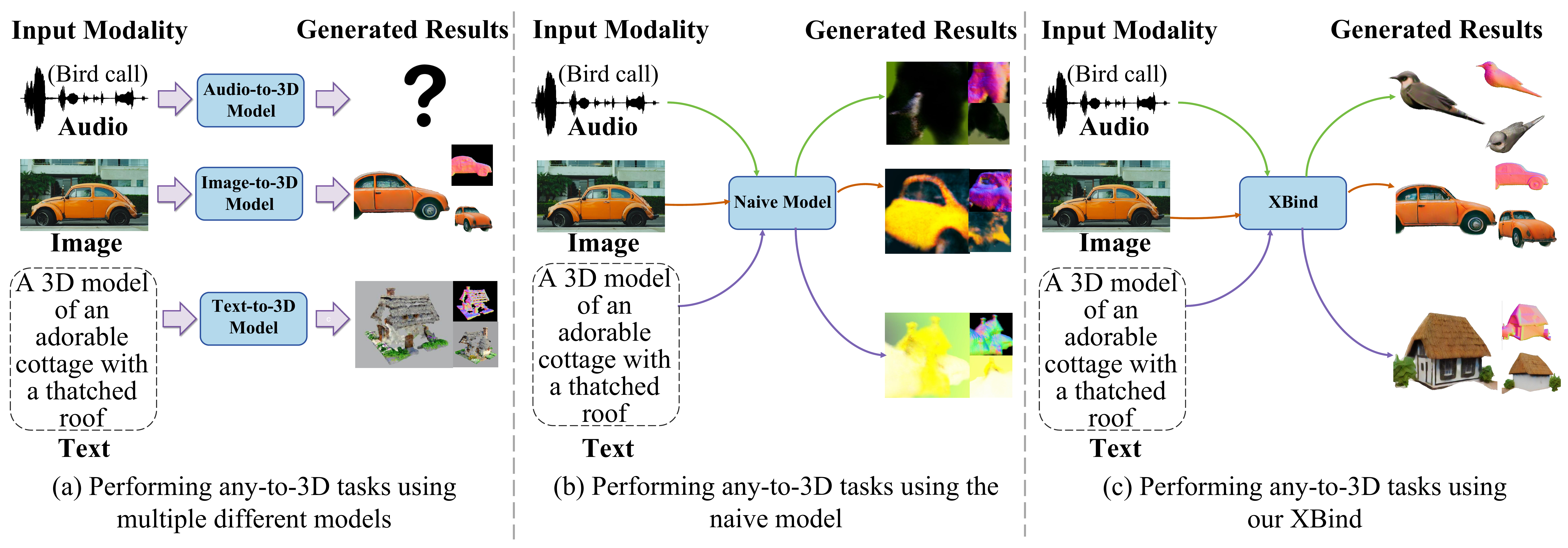} 
\caption{Comparison of various methods for Any-to-3D generation: (a) Utilizing separate expert models for Any-to-3D generation. (b) Simply concatenating \encodername and a 2D diffusion model to achieve Any-to-3D generation. (c) Our proposed \modelname. Since there are no existing audio-to-3D models, the audio prompt generated results in (a) are replaced with a question mark.}
\label{2d-to-3d failure case}
\end{figure*}

\IEEEPARstart{3}{D} content is crucial for enhancing visualization, understanding, and interaction with complex objects due to its resemblance to real-world environments~\cite{liu2024comprehensive}. However, creating high-quality 3D content is challenging, requiring specialized software and 3D modeling expertise~\cite{shi2023mvdream,sun2023dreamcraft3d,wang2024crm}.

Recently, diffusion models have garnered considerable attention in the field of image synthesis due to their remarkable capabilities, which can be attributed to the use of large-scale image-text datasets and scalable generative model architectures~\cite{peebles2023scalable,ronneberger2015u}. This success has naturally extended to 3D generation, where pre-trained 2D diffusion models~\cite{ramesh2022hierarchical,rombach2022high} have been adapted to facilitate the 3D content creation process~\cite{chen2023fantasia3d,wang2023score}. By automating the creation of 3D content, these advanced methodologies hold great promise for simplifying and enhancing the 3D content creation process.

Current 3D content generation predominantly relies on single modalities, such as text-to-3D~\cite{wu2024consistent3d,alldieck2024score,wang2024esd,wang2024prolificdreamer,michel2022text2mesh} or image-to-3D~\cite{wan2024cad,tang2023dreamgaussian,tang2023make,melas2023realfusion}, as shown in \figref{2d-to-3d failure case}-(a). This approach proves impractical for real-world applications where multiple modalities coexist, each contributing unique value. Generating 3D content separately for different modalities is both time-consuming and labor-intensive, undermining the automation of 3D content creation. Additionally, certain modalities, like audio, cannot be directly translated into 3D content, as shown in \figref{2d-to-3d failure case}-(a); transforming them into text or images first leads to information loss. Hence, a unified framework that accurately encapsulates the multimodal nature of reality is essential for achieving end-to-end multimodal 3D generation.

Research on multimodal-aligned encoders~\cite{girdhar2023imagebind,tang2024any} enables multiple modalities to be projected into a shared space, facilitating alignment. In the 2D domain, leveraging these aligned modalities has led to successful multimodal conditional content generation with 2D diffusion models~\cite{tang2024any}. 
However, when aligned modalities are combined with the mainstream single-modality 3D generation paradigm, which is based on 2D diffusion models~\cite{poole2022dreamfusion} for multimodal 3D synthesis, they often yield suboptimal results, as shown in \figref{2d-to-3d failure case}-(b). The reason may be the significant domain gap between 2D and 3D multimodal generation. Therefore, techniques designed for single modalities, such as text-to-3D and image-to-3D, cannot be directly transferred to multimodal 3D generation.
\IEEEpubidadjcol

In this paper, we introduce \modelname, a pioneering framework designed to unify multiple modalities for any modality conditional 3D generation. This framework overcomes the limitations of single-modality-based 3D generation and fills a gap in the any-to-3D field. Specifically, \modelname utilizes a \encodername~\cite{girdhar2023imagebind} to encode various modalities into a unified shared space. The resulting embeddings serve as prompts for the pretrained 2D diffusion model~\cite{ramesh2022hierarchical,rombach2022high}, with distillation sampling loss~\cite{poole2022dreamfusion,wu2024consistent3d} guiding the optimization of 3D representations.
To mitigate the issue of inadequate expression of input modalities in generated 3D objects (\figref{2d-to-3d failure case}), our framework introduces a novel loss function, \fullnamelossb. This function aligns the embeddings of any modality with the embeddings of images rendered from 3D objects at different camera viewpoints, thereby enhancing the incorporation of modality information into the 3D optimization process. By leveraging \nameloss, \modelname produces 3D objects that more precisely adhere to the prompts of different modalities.
Additionally, to address the severe Janus problem associated with using only the 2D diffusion model and \nameloss as pixel-level planar supervision for guiding 3D object generation, we incorporate a 3D-aware diffusion model~\cite{liu2023zero} as a spatial-level stereoscopic supervision. This supervision, combined with pixel-level planar supervision, forms a Hybrid Diffusion Supervision that guides the generation of 3D objects, ensuring high-quality, view-consistent any-to-3D generation.
\modelname employs a Three-Phase optimization approach that progresses from coarse to fine detail. In the first phase, a low-resolution Neural Radiance Field (\nerf)~\cite{barron2021mip,muller2022instant} is optimized. The optimized \nerf then initializes a high-resolution \dmtet~\cite{shen2021deep}. The second and third phases focus on optimizing the geometry and texture of the \dmtet, respectively, culminating in high-fidelity 3D object generation.
We conducted extensive experiments to evaluate the effectiveness of \modelname. The results demonstrate that \modelname is capable of generating high-quality 3D objects that are well-aligned with the given modality prompts. As the first unified framework for any-to-3D generation, \modelname significantly reduces both time and resource consumption by enabling the direct generation of 3D objects from any user-provided modalities. Additionally, it mitigates information loss associated with modality transformation.

Our main contributions are summarized as follows:
\begin{itemize}
    \item We introduce \fullnamelossb, a novel loss function designed to improve 3D generation results by effectively combining the \encodername and the 2D diffusion model.
    \item We present a coarse-to-fine Three-Phase framework that leverages Hybrid Diffusion Supervision, significantly enhancing the visual quality and consistency of any-to-3D generation.
    \item Building on these two innovations, we introduce \modelname, a pioneering unified framework for any-to-3D generation. Extensive experiments validate the superior performance of \modelname.
\end{itemize}

\section{Related Works}
\subsection{Diffusion Models}
Diffusion models (DMs) learn data distributions through denoising and recovering original data~\cite{yang2023diffusion,song2020score,ho2020denoising}. Enhanced by pre-trained models~\cite{radford2021learning}, diffusion models have shown significant advancements in image~\cite{gu2022vector,dhariwal2021diffusion,yang2024improving,zhou2023shifted}, video~\cite{khachatryan2023text2video,luo2023videofusion,mei2023vidm,wang2023lavie}, speech~\cite{huang2022prodiff,zhu2023taming,richter2023speech,lemercier2023storm}, and 3D synthesis~\cite{babu2023hyperfields,cao2023texfusion,liang2024luciddreamer,chen2024scenetex}, revolutionizing the field of computer vision. 
Pioneering works like Stable Diffusion~\cite{rombach2022high} and DeepFloyd generate high-quality images from text prompts by learning priors from large-scale datasets. By leveraging large-scale image datasets, these diffusion models learn various priors ranging from object appearance to complex scene layouts. A series of subsequent works have fine-tuned text-to-image diffusion models, successfully extending their capabilities to better adapt to different downstream tasks. For instance, Stable UnCLIP~\cite{Rombach_2022_CVPR} can accept CLIP image embeddings in addition to text prompts and can be used to create image variations or be combined with text-to-image CLIP priors in a chained manner. Additionally, works like~\cite{liu2023zero,liu2023syncdreamer} learn 3D-aware diffusion models by rendering images from synthetic objects~\cite{deitke2023objaverse,deitke2024objaverse}. Diffusion models are powerful tools for complex data modeling and generation. Their robust and stable capabilities in handling complex data have led to their successful application across various domains.

\subsection{Multimodal Representation Learning}
The field of multimodal modeling has experienced rapid development recently. Some studies have explored the joint training of multiple modalities in both supervised~\cite{girdhar2022omnivore,likhosherstov2021polyvit} and self-supervised contexts~\cite{arandjelovic2017look,girdhar2023omnimae,morgado2021audio,tian2020contrastive,wang2022bevt}, aiming to build unified representations of various modalities using a single model to achieve a more comprehensive cross-modal understanding. The Vision Transformer (ViT)~\cite{dosovitskiy2020image}, due to its diverse model architectures and training methods, has been widely applied to downstream tasks such as visual question answering and image captioning. Multimodal encoders have also achieved significant success in the domains of vision-language~\cite{alayrac2022flamingo,cho2021unifying,zellers2021merlot,tsimpoukelli2021multimodal,yang2023vilam}, video-audio~\cite{tang2022tvlt}, and video-speech-language~\cite{yang2023code,zellers2022merlot}. Aligning data from different modalities is an active area of research~\cite{girdhar2023imagebind,tang2024any,tang2024codi,elizalde2023clap,radford2021learning}, showing great potential for applications in cross-modal retrieval and building unified multimodal representations~\cite{liu2023audioldm,mokady2021clipcap,rombach2022high}.

\begin{figure*}[t]
\centering
\includegraphics[width=0.9\textwidth]{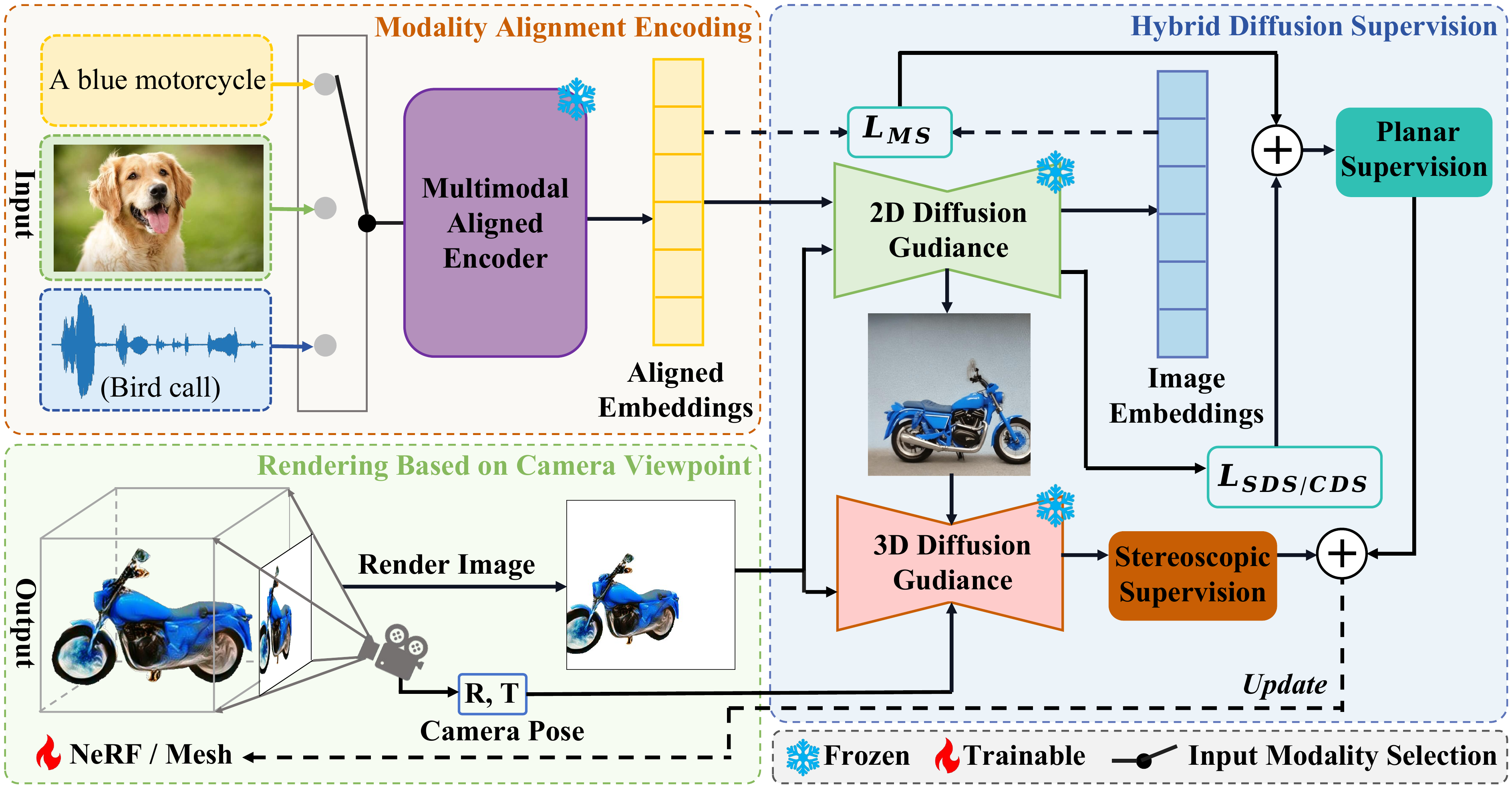} 
\caption{Overview of our method. \modelname first encodes the input modality using a \encodername, mapping it into a shared modality space. This aligned modality is then used as a condition for both 2D and 3D diffusion models. Hybrid diffusion supervision, combining planar and stereoscopic supervision, is applied to optimize the \nerf/Mesh.}
\label{pipeline}
\end{figure*}

\section{Preliminaries}

\subsection{Consistency Distillation Sampling}
Consistency Distillation Sampling (CDS) explores the deterministic sampling prior of the ordinary differential equation (ODE) for 3D generation~\cite{wu2024consistent3d}, addressing the uncertainties associated with the stochastic differential equation (SDE) corresponding to the existing and popular Score Distillation Sampling (SDS)~\cite{poole2022dreamfusion}. This approach mitigates the issues of geometry collapse and poor textures in 3D objects that arise due to the uncertainty in SDE, thereby enabling the generation of high-fidelity and diverse 3D objects. 

Specifically, during each training iteration, given an image rendered by a 3D model, the target 3D score function is first estimated using a pre-trained 2D diffusion model and an ODE is constructed for trajectory sampling accordingly. Two adjacent samples are then sampled along the ODE trajectory, using the less noisy sample to guide the more noisy one, thereby distilling the deterministic prior into the 3D model. 

The CDS loss effectively addresses the challenges associated with the randomness in the solution distribution of the SDE corresponding to the SDS loss, providing a more reliable and consistent framework to guide the text-to-3D generation process. In this work, we employ CDS as the supervisory loss for the 2D diffusion model in the geometry refinement phase.

\subsection{\imagebind}
\imagebind~\cite{girdhar2023imagebind} is a method that learns a shared representation space by leveraging various types of image-paired data. It does not require datasets where all modalities co-occur; instead, image-paired data alone is sufficient to bind these modalities together. \imagebind can utilize recent large-scale vision-language models~\cite{radford2021learning} and extend their zero-shot capabilities to new modalities through the natural pairing of other modalities with images. Specifically, consider a pair of modalities $(\mathcal{I,M} )$ with aligned observations, where $\mathcal{I}$ represents images and $\mathcal{M}$ is another modality. Given an image $\mathrm{I}_{i}$ and its corresponding observation in the other modality $\mathrm{M}_{i}$, \imagebind encodes them into normalized embeddings: $\mathrm{x}_{i}=f(\mathrm{I}_{i})$ and $\mathrm{y}_{i}=g(\mathrm{M}_{i})$, where $f$ and $g$ are deep networks. The embeddings and encoders are optimized using the InfoNCE~\cite{oord2018representation} loss:

\begin{equation}
\mathcal{L}_{(\mathcal{I,M} )} =-log\frac{exp(\mathrm{x}_{i}^{\mathrm{T}}\mathrm{y} _{i}/t)}{exp(\mathrm{x}_{i}^{\mathrm{T}}\mathrm{y} _{i}/t)+ {\textstyle \sum_{j\ne i}^{}exp(\mathrm{x}_{i}^{\mathrm{T}}\mathrm{y} _{j}/t)} }, 
\end{equation}

where $t$ is a scalar temperature that controls the smoothness of the softmax distribution, and $j$ denotes unrelated observations, also known as "negative samples." This loss function brings the embeddings $\mathrm{x}_{i}$ and $\mathrm{y} _{i}$ closer in the joint embedding space, thereby aligning $\mathcal{I}$ and $\mathcal{M}$. Simultaneously, an emergent behavior appears in the embedding space, which aligns the modality pairs $\mathcal{M}_{1} $ and $\mathcal{M}_{2} $ even when trained only with the modality pairs $(\mathcal{I,M}_{1} )$ and $(\mathcal{I,M}_{2} )$. This behavior enables \imagebind to perform rich compositional multimodal tasks across different modalities. However, these tasks have so far been limited to the 2D domain, and how to apply \imagebind to the 3D domain remains unexplored. In this work, we utilize \imagebind as the \encodername for \modelname to achieve 3D generation from multiple modalities.

\section{Method}
In this section, we present our framework, \modelname,  designed to handle any modality inputs and generate high-quality 3D objects based on these inputs. Our framework utilizes a coarse-to-fine strategy, dividing the model optimization into three phases. As illustrated in \figref{pipeline}, we provide a clear design space for optimization at each phase.
The first phase involves coarse 3D object generation using \nerf as the 3D representation. \modelname encodes input from any modality into a shared latent space, which is then used to condition both 2D and 3D diffusion models. The model then leverages hybrid diffusion supervision to optimize the \nerf, generating a coarse but consistent 3D object.
In the second phase, geometry optimization, the model extracts the \nerf representation as a \dmtet and refines the geometric details of the 3D mesh using the same supervision.
The third phase focuses on texture optimization, where the model enhances texture details with hybrid diffusion supervision. After completing these phases, \modelname produces high-quality 3D objects with consistent and rich geometric and textural details.

\subsection{\fullnamelossb}
When using embeddings from the \encodername as prompts for the 2D diffusion model to guide 3D generation, the results are poor and exhibit mode collapse, as shown in \figref{2d-to-3d failure case}-(b). This issue likely arises because the direct guidance of modality embeddings during 3D generation exacerbates the gap between 2D and 3D domains.

To address this challenge, we propose a \fullnamelossb. By aligning the embeddings encoded by the \encodername with the CLIP embeddings of the images rendered from 3D objects, this approach enhances the correct guidance of various modalities in the 3D object generation process. Instead of treating the images rendered from different camera viewpoints merely as latents for denoising, we also encode them into the diffusion model’s conditional space via the CLIP image encoder. Specifically, we use a differentiable renderer to render a set of images $\mathbf{x}$ from the corresponding \nerf or Mesh under a given set of camera viewpoints $\mathrm{p}$. Subsequently, data augmentation is applied to $\mathbf{x}$ to obtain fine-grained geometry and texture. Data augmentation includes global augmentation $G(\cdot)$, local augmentation $L(\cdot)$, and normalization $Z(\cdot)$. $G(\cdot)$ applies random perspective transformations to rendered images, while $L(\cdot)$ involves random cropping and perspective transformations. The augmented rendered images are then fed into the CLIP image encoder~\cite{radford2021learning} to obtain their embeddings. Subsequently, we compute the \nameloss between the embedding of the input modality prompt and those of the rendered images as follows:
\begin{equation}
\begin{split}
&\mathcal{L}_{\mathit{MS}}=\omega_{g}\mathcal{L}_{\mathit{ms\text{-}g}}+\omega_{l}\mathcal{L}_{\mathit{ms\text{-}l}}+\omega_{z}\mathcal{L}_{\mathit{ms\text{-}z}},\\
&\mathcal{L}_{\mathit{ms\text{-}g}}=  \sum_{i}^{n}  \left \{   W_{i}\times \mathrm{avg}    \left[ cos \left (\mathbf{\mathit{C}}_m,  \mathcal{E} (G(\mathrm{x})) \right) \right ]\right \} ,\\
&\mathcal{L}_{\mathit{ms\text{-}l}}= \sum_{i}^{n}  \left \{   W_{i}\times \mathrm{avg}    \left[ cos \left (\mathbf{\mathit{C}}_m,  \mathcal{E} (L(\mathrm{x})) \right) \right ]\right \} ,\\
&\mathcal{L}_{\mathit{ms\text{-}z}}= \sum_{i}^{n}  \left \{   W_{i}\times \mathrm{avg }   \left[ cos \left (\mathbf{\mathit{C}}_m,  \mathcal{E} (Z(\mathrm{x})) \right) \right ]\right \} ,
\end{split}
\label{msloss}
\end{equation}
where $n$ represents the number of data augmentations in each iteration, $W_{i}$ is the weight of the similarity during the $i\text{-th}$ augmentation, $\mathrm{avg} [\cdot ]$ denotes the calculation of the average similarity between the embedding of the input modality and the embeddings of all rendered images, $cos(a,b)$ is the cosine similarity between $a$ and $b$,
$\mathbf{\mathit{C}}_m$ represents the embedding of the modality prompt encoded by the \encodername, $\mathcal{E}$ denotes the CLIP image encoder, and $\omega_{g}$, $\omega_{l}$, $\omega_{z}$ are the weighting parameters. 

By applying the \nameloss, the semantic information of the modality prompts can more accurately guide the optimization process of the 3D objects, thereby enhancing the supervisory effect of the 2D image diffusion model.

\subsection{Hybrid Diffusion Supervision}
\subsubsection{Pixel-Level Planar Supervision}
To enhance the diversity of 3D objects generated by \modelname, we employ 2D diffusion model\footnote{Our 2D diffusion model is based on Stable Diffusion v2-1-unclip, which is a fine-tuned version of Stable Diffusion 2.1, modified to accept CLIP image embeddings.} and \nameloss together as pixel-level planar supervision to guide \modelname during the 3D object generation process. The corresponding supervision losses are discussed as follows:

\paragraph{Consistency Distillation Sampling Loss.} We employ supervision with CDS loss~\cite{wu2024consistent3d} to achieve fine geometry in the 3D generated objects. As previously discussed, the CDS method employs ODE deterministic sampling priors for 3D generation. The specific formulation of the CDS loss $\mathcal{L}_{CDS}$ is as follows:
\begin{equation}
\begin{split}
    &\mathbb{E}_{\mathrm{p} }\big[\lambda(t_{2} ){||D_{\phi}(\mathrm{z} _{t_{1}},t_{1},\mathbf{\mathit{C}}_m)-\mathrm{sg} (D_{\phi}(\hat{\mathrm{z}} _{t_{2}} ,t_{2},\mathbf{\mathit{C}}_m))||}_{2}^{2} \big],
    \\
    &\mathrm{z}_{t_{1}} = \mathrm{z}_{p} + \sigma_{t_{1}}\epsilon^{*},\\ 
    &\mathrm{\hat{z}}_{t_{2}} = \mathrm{z}_{t_{1}} + \frac{\sigma _{t_{2}} - \sigma _{t_{1}}}{\sigma _{t_{1}}}(\mathrm{z}_{t_{1}} - D_{\phi}(\mathrm{z}_{t_{1}},t_{1},\mathbf{\mathit{C}}_m)),
\end{split}
\end{equation}
where $\lambda(t_{2} )$ denotes the loss weight, $D_{\phi}(\cdot )$ is a pre-trained 2D diffusion model, $\mathrm{sg} (\cdot )$ is a stop-gradient operator, $\mathrm{z}_{t_{1}}$denotes the noisy latent vector, $\mathrm{z}_{p}$ represents the latent vector of the image of a 3D object rendered from camera view $p$, $\sigma_{t}$ varies along time-step $t$, $\epsilon^{*}$ is a fixed Gaussian noise, $ t_{1}>t_{2}$ are two adjacent diffusion time steps and $\hat{\mathrm{z}} _{t_{2}}$ is a less noisy latent vector derived from deterministic sampling by running one discretization step of a numerical ODE solver from $\mathrm{z}_{t_{1}}$.

We found that using the CDS loss can improve the geometry quality of the 3D objects, capturing more details.

\paragraph{Augmented 2D SDS Loss.} While CDS loss can refine geometry, it fails to generate correct geometry and textures when applied during the initial generation, as shown in the second column of the~\figref{comparison}. To address this issue, we employed the Score Distillation Sampling (SDS) loss from Dreamfusion~\cite{poole2022dreamfusion}. SDS distills the 2D priors of a pre-trained diffusion model into a 3D model, parameterized by $\theta$ (\emph{e.g.,} NeRF, Mesh). Specifically, given an image $\mathrm{x} =g(\theta)$ of a 3D object rendered from a specific camera view, where $g(\cdot )$ is a differentiable renderer, SDS utilizes the 2D diffusion model to encode it into latent variables $\mathrm{z}$ and adds noise to obtain $\mathrm{z}_\mathrm{t}$. Under the guidance of the corresponding modality prompt input $y$, SDS applies a denoising training objective to the noisy rendered image, predicting a clear novel view. The gradient for the SDS loss is computed as follows:
\begin{equation}
\begin{split}
    &\bigtriangledown _{\theta }\mathcal{L}_{SDS}(\phi ,\,\mathrm{z}) =\mathbb{E}_{t,\epsilon } \left [\lambda (t)(\hat{\epsilon}_{\phi}(\mathrm{z}_\mathrm{t};\mathbf{\mathit{C}}_m,t) -\epsilon )\frac{\partial\mathrm{z}}{\partial \theta}\right ]\\
    & =\mathbb{E}_{t,\epsilon }\left [\lambda (t)(\mathrm{z}- \hat{\mathrm{z}} )\frac{\partial\mathrm{z}}{\partial \theta}\right ],
\end{split}
\label{origin_sds}
\end{equation}
where $\phi$ represents the parameters of the pre-trained 2D diffusion model, $\hat{\epsilon}_{\phi}$ denotes the predicted noise, $\epsilon$ stands for the standard Gaussian noise, and $\hat{\mathrm{z}}$ represents the estimate of the latent vector $\mathrm{z}$ using the denoising function $\hat{\epsilon}_{\phi}$. The term $(\mathrm{z}- \hat{\mathrm{z}} )$ is referred to as the latent vector residual.

Building upon this foundation, to achieve higher fidelity in the 3D objects generated by \modelname, we extend the SDS loss to the image space, following~\cite{zhu2023hifa}. The gradient representation of the loss function changes from Eq.~\ref{origin_sds} to the following form:
\begin{equation}
\begin{split}
    &\bigtriangledown _{\theta }\mathcal{L}_{SDS^{*}}(\phi ,\mathrm{z},\mathrm{x} =g(\theta )) \\
    &=\mathbb{E}_{t,\epsilon }\left \{ \lambda (t)\left [ ( \mathrm{z}-\hat{\mathrm{z}})\frac{\partial\mathrm{z}}{\partial \theta}+\omega_{img} (\mathrm{x}- \hat{\mathrm{x}} )\frac{\partial\mathrm{x}}{\partial \theta}\right ]\right \},
\end{split}
\end{equation}
where $\hat{\mathrm{x}}$ is the estimated image obtained by decoding the latent vector $\hat{\mathrm{z}}$ predicted by the diffusion model, and $\omega_{img}$ is a weighting parameter.

Simultaneously, to mitigate the issue of geometric blur in \nerf and provide a well-initialized 3D object for the subsequent geometry refinement phase, we follow~\cite{zhu2023hifa} and adopt a regularization method that minimizes the variance of sampled z-coordinates distributed along \nerf rays. This loss is incorporated as part of the augmented 2D SDS loss. A smaller variance indicates a clearer geometric surface. The variance $\sigma_{z}^{2}$ of the z-coordinates along a ray $r$ can be expressed as follows:
\begin{equation}
\begin{split}
    &\sigma_{z}^{2} =\mathbb{E}_{z}\left [ (z_{i}-d_{z})^{2} \right ] = {\textstyle \sum_{i}^{n}(z_{i}-d_{z})^{2}\frac{\mathrm{w} _{i}}{\sum _{i}\mathrm{w} _{i}} }\:\mathrm{with}\\
    & d_{z}=\textstyle \sum_{i}^{n}z_{i}\frac{\mathrm{w} _{i}}{\sum _{i}\mathrm{w} _{i}},
\end{split}
\end{equation}
where $z_{i}$ represents the z-coordinate sampled along the ray $r$, $n$ is the number of sampling points, $d_{z}$ denotes the depth value of the ray, and $\mathrm{w} _{i}$ represents the weights of the sampled $z_{i}$.

The regularization loss $\mathcal{L}_{nz}$ for the variance $\sigma_{z}^{2}$ is given by the following formula:
\begin{equation}
\begin{split}
    \mathcal{L}_{nz} =\mathbb{E}\left [ \delta \,\sigma_{z}^{2} \right ] \:\mathrm{with}\:\delta=\begin{cases}
 1\qquad\mathrm{if}\: \textstyle\sum_{i}^{n}\mathrm{w}_{i}>0.5\\
0\qquad\mathrm{if}\: \textstyle\sum_{i}^{n}\mathrm{w}_{i}\le 0.5
\end{cases},
\end{split}
\end{equation}
where $\delta$ functions as an indicator or binary weight to exclude background rays.

Thus, the augmented 2D SDS loss is defined as:
\begin{equation}
    \mathcal{L}_{Aug-SDS}=\omega^{*}\mathcal{L}_{SDS^{*}}+\omega_{nz}\mathcal{L}_{nz}.
\label{augsds}
\end{equation}
where $\omega^{*}$ and $\omega_{nz}$ are the weighting parameters for the respective loss functions. Specifically, we use $\mathcal{L}_{nz}$ only when optimizing the \nerf.

In summary, the complete pixel-level planar supervision is composed of the aforementioned \nameloss, the CDS loss, and the augmented 2D SDS loss.  The CDS loss and the augmented 2D SDS loss are applied at different phases of \modelname optimization. The specific formula for pixel-level planar supervision is given as follows:
\begin{equation}
    \mathcal{L}_{planar}=\omega_{ms}\mathcal{L}_{MS}+\left [ \omega_{cds}\mathcal{L}_{CDS},\mathcal{L}_{Aug-SDS} \right ], 
    \label{planar}
\end{equation}
where $\omega_{ms}$ and $\omega_{cds}$ are the weighting parameters for the respective loss functions, and $[\, , ]$ indicates that only one of the components is used as supervision at different phases of \modelname optimization.

Pixel-level planar supervision has demonstrated exceptional generalization capabilities in generating 3D objects with \modelname. It enhances 3D imagination, possesses significant advantages in exploring geometric spaces, and promotes realistic appearances and reasonable scene layouts. This is because the 2D diffusion models it relies on are trained on extensive datasets containing billions of images. However, solely relying on pixel-level planar supervision may introduce inaccuracies in the generated 3D representations due to insufficient 3D knowledge. This inevitably leads to deficiencies in 3D fidelity and consistency. Such shortcomings can result in unrealistic geometric structures, such as the problem of multiple faces (Janus problem).

\subsubsection{Spatial-Level Stereoscopic Supervision}
To overcome the limitations of using only 2D diffusion models for guiding the generation of 3D objects, we introduce Zero-1-to-3~\cite{liu2023zero} as a 3D-aware diffusion model, to work in conjunction with the 2D diffusion model. By integrating both models, the generated 3D objects exhibit superior and consistent geometry and texture. The incorporation of Zero-1-to-3 also introduces several different types of supervision, which are discussed in detail below.


\paragraph{3D SDS Loss.} The generation of 3D objects is supervised using the 3D SDS loss. The specific gradient of the 3D SDS loss is as follows:
\begin{equation}
\bigtriangledown_{\theta }\mathcal{L}_{3D-SDS}=\mathbb{E}_{t,\epsilon }[\lambda (t)(\hat{\epsilon}_{\phi}(\mathrm{z}_\mathrm{t};c(\mathrm{x}_{r},R,T),t) -\epsilon )\frac{\partial\mathrm{z}}{\partial \theta}],
\end{equation}
where $\hat{\epsilon}_{\phi}$ represents a denoising function, $\epsilon$ is standard Gaussian noise, $\mathrm{x}_{r}$ denotes the reference view\footnote{To mitigate information loss from direct modality conversion, \modelname uses \encodername to map user-provided modalities into a unified space. The aligned embeddings then guide the 2D diffusion model to generate images encapsulating the modality information, which serve as conditional input for the 3D-aware diffusion model.}, $R\in \mathbb{R} ^{3\times 3}$ and $T\in \mathbb{R} ^{3}$ represent the relative camera rotation and translation of the desired viewpoint, respectively, and $c(\mathrm{x}_{r},R,T)$ is the embedding of the reference view and relative camera extrinsic parameters.

\paragraph{Reference View Loss.} To maximize the guidance potential of the 3D-aware diffusion model, we introduce the reference view loss. This loss ensures the quality of the 3D object generated from the reference viewpoint. The specific form of the loss function is as follows:
\begin{equation}
    \mathcal{L}_{ref}=\omega_{rgb}\left \| \mathrm{m}\odot(\mathrm{x}_{r}-\hat{ \mathrm{x}}_{r} )\right \|_{2}^{2}+\omega_{mask}\left \| \mathrm{m}-M(\hat{ \mathrm{x}}_{r} ))\right \|_{2}^{2},
\end{equation}
where $\mathrm{m}$ is the reference view mask,
$\hat{ \mathrm{x}}_{r} $ is the image rendered from the reference viewpoint, $\odot$ is Hadamard product, and $M(\cdot)$ is the foreground mask. 

Together, these two types of supervision constitute the spatial-level stereoscopic supervision. The specific form of the loss function is as follows:

\begin{equation}
    \mathcal{L}_{stereo}=\omega_{3d}\mathcal{L}_{3D-SDS}+\mathcal{L}_{ref}.
\label{stereo}
\end{equation}

Under the influence of these two types of supervision, \modelname incorporates the prior knowledge from the 3D-aware diffusion model into the generation of 3D objects. This effectively addresses the issues of 3D fidelity and consistency that arise when using only 2D diffusion models for guidance.

In summary, pixel-level planar supervision and spatial-level stereoscopic supervision together form the Hybrid Diffusion Supervision in \modelname's three-phase optimization process, as shown in the following formula:

\begin{equation}
    \mathcal{L}_{hybrid}=\mathcal{L}_{planar}+\mathcal{L}_{stereo}.
\end{equation}

Pixel-level planar supervision enhances the imaginative and generalization capabilities of 3D generation but struggles with 3D inconsistency and uncertainty. In contrast, spatial-level stereoscopic supervision ensures more accurate 3D geometry and consistency but lacks generalizability due to the limited scale of 3D training datasets, often leading to overly smooth surfaces. By combining these two types of supervision as the  hybrid diffusion supervision, \modelname effectively harnesses their strengths and compensates for their weaknesses, enabling the generation of realistic, detailed, and 
consistent 3D objects.

\begin{algorithm}[tb]
\caption{Three-Phase Optimization}
\label{algorithm}
\begin{algorithmic}[1]
\Require
\Statex $\mathrm{NR}_{\theta}$ \Comment{Neural Radiance Fields}
\Statex $\mathrm{DM}_{\theta}$ \Comment{\dmtet hybrid scene representation}
\Statex $\eta$ \Comment{learning rate}
\Statex $N_{phase\text{-1}}, N_{phase\text{-2}}, N_{phase\text{-3}}$ \Comment{iterations for each phase}
\Statex $\mathcal{L}_{planar\text{-}S},\mathcal{L}_{planar\text{-}C},\mathcal{L}_{stereo}$ \Comment{hybrid supervision}
\Statex $\mathcal{L}_{nv},\mathcal{L}_{nc},\mathcal{L}_{ls}$ \Comment{geometric regularization losses}
\Statex
\State * Phase I
\For{iter in $N_{phase\text{-1}}$} \Comment{\nerf update}
    \State $\mathcal{L}_{hybrid}\gets \mathcal{L}_{planar\text{-}S}+\mathcal{L}_{stereo}$
    \State $Loss\gets \mathcal{L}_{hybrid}+\mathcal{L}_{nv}$
\EndFor
\Statex
\State * Phase II
\State $\mathrm{NR}_{\theta}\Rightarrow \mathrm{DM}_{\theta}$\Comment{\dmtet-geometric update}
\For{iter in $N_{phase\text{-2}}$} 
    \State $\mathcal{L}_{hybrid}\gets \mathcal{L}_{planar\text{-}C}+\mathcal{L}_{stereo}$
    \State $Loss\gets \mathcal{L}_{hybrid}+\mathcal{L}_{nc}+\mathcal{L}_{ls}$
\EndFor
\Statex
\State * Phase III
\State freeze the geometric shape parameters of the $\mathrm{DM}_{\theta}$
\For{iter in $N_{phase\text{-3}}$} \Comment{\dmtet-texture update}
    \State $\mathcal{L}_{hybrid}\gets \mathcal{L}_{planar\text{-}S}+\mathcal{L}_{stereo}$
    \State $Loss\gets \mathcal{L}_{hybrid}$
\EndFor
\Statex
\Procedure{UPDATE}{Loss}
    \State $\mathrm{x}=\begin{cases}
  \mathrm{NR}_{\theta}\to \text{rgb img,}& \text{ phase 1}\\
  \mathrm{DM}_{\theta}\to\text{normal map,}& \text{ phase 2 }\\
  \mathrm{DM}_{\theta}\to \text{rgb img,}& \text{ phase 3 }
\end{cases}$ \Comment{\parbox[t]{1.2cm}{render\\images}}
    \State $\theta \gets \theta-\eta \bigtriangledown _{\theta }(Loss)$ \Comment{optimize $\mathrm{NR}_{\theta}$ or $\mathrm{DM}_{\theta}$}
\EndProcedure
\end{algorithmic}
\end{algorithm}
\subsection{Three-Phase Optimization}
To achieve high-fidelity 3D object generation, we propose a three-phase optimization method. By integrating each phase with our proposed hybrid diffusion supervision, \modelname can progressively optimize 3D objects from coarse to fine, resulting in high-quality geometry and texture.

\subsubsection{Phase I - Coarse Optimization}
The optimization objective of the first phase is to learn a coarse texture and 3D geometric shape that aligns with the modality prompt. Given that \nerf excels in smoothly handling complex topological changes, we adopt a low-resolution implicit \nerf for 3D representation in this phase. The \nameloss and the augmented 2D SDS loss from pixel-level planar supervision ($\mathcal{L}_{planar}$) are used in this phase, along with spatial-level stereoscopic supervision ($\mathcal{L}_{stereo}$) and normal vector regularization ($\mathcal{L}_{nv}$)~\cite{melas2023realfusion} to update the \nerf until convergence.


\subsubsection{Phase II - Geometric Refinement}
To enhance the geometric details of the 3D objects represented by the low-resolution \nerf obtained in the first phase, we convert the neural field into a Signed Distance Field (SDF) by subtracting a fixed threshold and employ high-resolution \dmtet~\cite{shen2021deep} as the 3D representation for subsequent optimization. The optimization objective of this phase is to refine the geometry of the 3D objects. Therefore, we follow~\cite{chen2023fantasia3d} and use the rendered normal maps of the 3D objects as inputs to the diffusion models. During this phase, we utilize the \nameloss and CDS loss from pixel-level planar supervision ($\mathcal{L}_{planar}$), and also employ spatial-level stereoscopic supervision ($\mathcal{L}_{stereo}$), normal consistency loss ($\mathcal{L}_{nc}$), and Laplacian smoothness loss ($\mathcal{L}_{ls}$)~\cite{wu2024consistent3d} to optimize the 3D representation. This results in 3D objects with refined surface geometry.

\subsubsection{Phase III - Texture Refinement}
In \textit{Phase II}, we transitioned the 3D representation from the implicit \nerf to the explicit \dmtet, allowing us to easily decouple the geometry and texture of 3D objects for separate optimization. The objective of this phase is to generate more detailed and rich textures for the 3D objects. During this phase, we input the rendered color images from \dmtet into both the 2D diffusion model and the 3D-aware diffusion model. By leveraging the \nameloss and the augmented 2D SDS loss from pixel-level planar supervision ($\mathcal{L}_{planar}$), as well as spatial-level stereoscopic supervision ($\mathcal{L}_{stereo}$), we systematically optimize the 3D representation to produce high-fidelity 3D objects that exhibit rich texture details.

The overall optimization process is illustrated in Algorithm~\ref{algorithm}, where $\mathcal{L}_{planar\text{-}C}$ denotes the pixel-level planar supervision using the CDS loss, $\mathcal{L}_{planar\text{-}S}$ denotes the pixel-level planar supervision using the augmented 2D SDS loss. By employing our proposed three-phase optimization method, \modelname effectively leverages the advantages of both 2D diffusion model and 3D-aware diffusion model. This approach thoroughly distills their prior knowledge into the 3D object optimization process, resulting in the generation of 3D objects with intricate geometry and texture details that align with the modality prompts.

\begin{figure*}[t]
\centering
\includegraphics[width=0.9\textwidth]{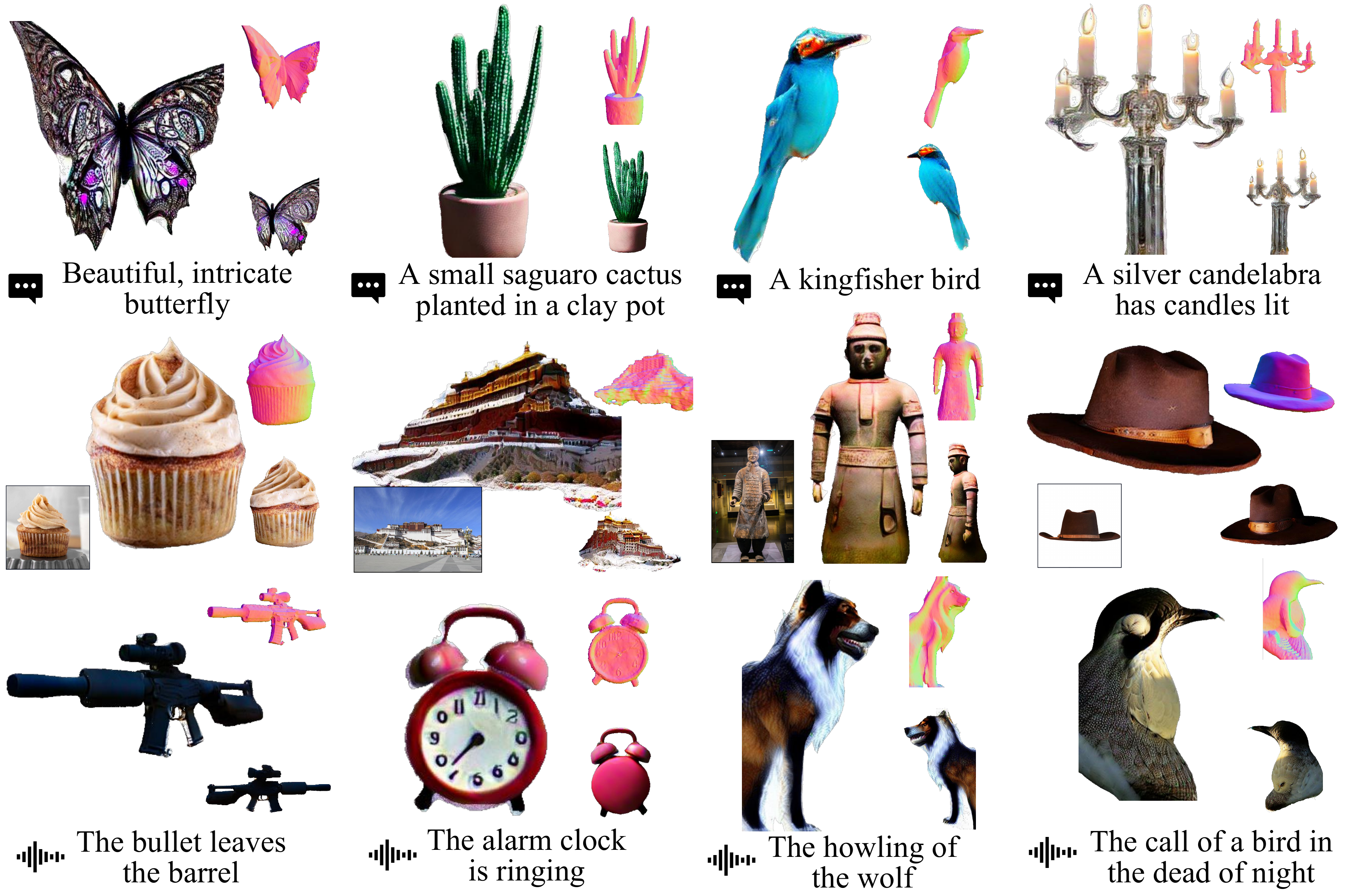} 
\caption{Examples generated by \modelname. The first row represents text-to-3D, the second row represents image-to-3D with the image prompt input located at the bottom left corner of each generated result, and the third row represents audio-to-3D.}
\label{examples}
\end{figure*}

\begin{figure*}[t]
\centering
\includegraphics[width=0.9\textwidth]{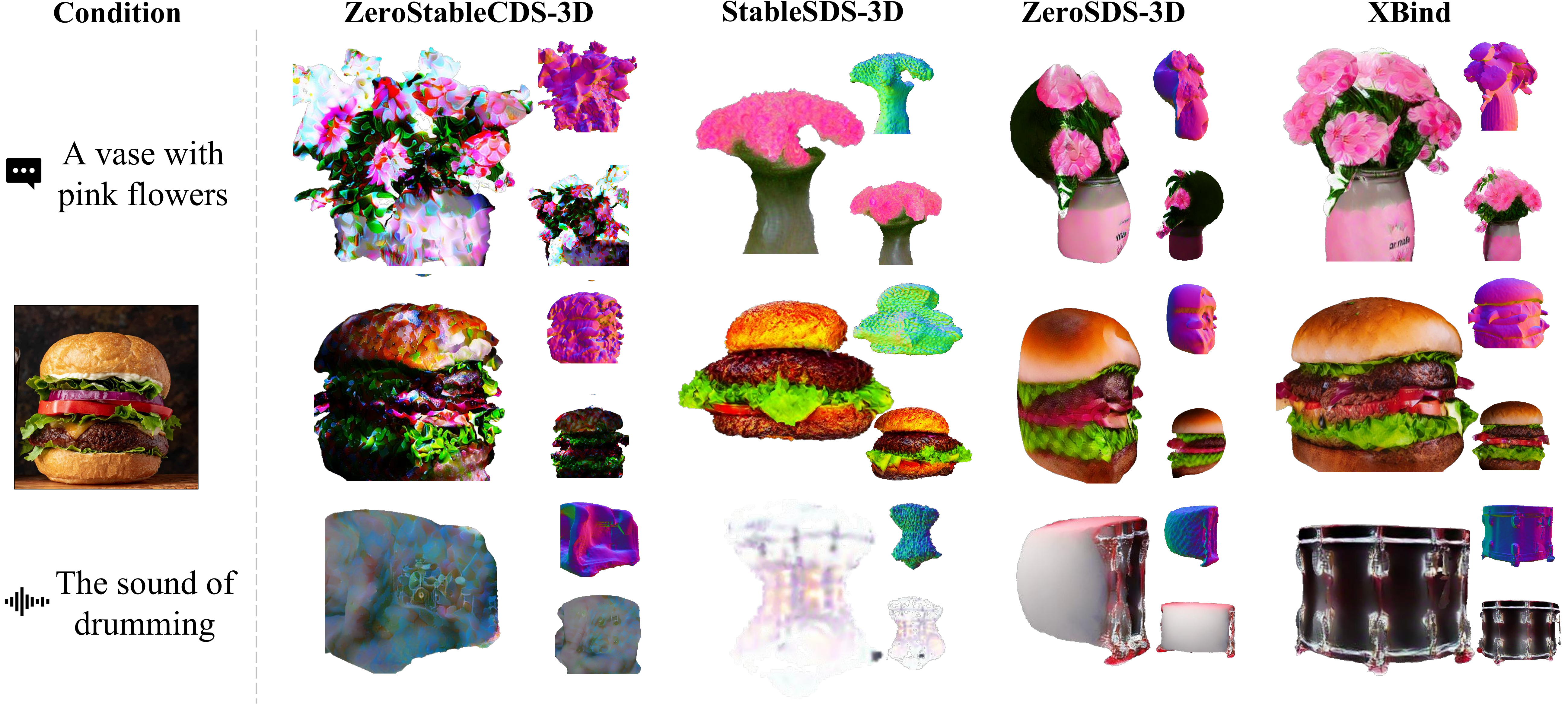} 
\caption{Qualitative comparison with baselines. The first row represents text-to-3D, the second row represents image-to-3D, and the third row represents audio-to-3D.}
\label{comparison}
\end{figure*}

\section{Experiment}
\subsection{Experimental Setup}
\subsubsection{Camera Settings}
The optimization process of \modelname is divided into three phases. In the first phase (the coarse optimization of \nerf), and the third phase (the texture refinement of \dmtet), the camera elevation angle is sampled between $-45$ and $45$ degrees, and the azimuth angle is sampled between $-180$ and $180$ degrees. During training in these two phases, the camera distance is set to $2.5$, and the vertical field of view (FoV) is set to $40$ degrees. In the second phase (the geometric refinement of \dmtet) of \modelname optimization, the camera elevation angle is randomly sampled from $-10$ to $45$ degrees, the azimuth angle from $-135$ to $225$ degrees, and the camera distance from $1.5$ to $2.0$. Additionally, in this phase, the vertical FoV is randomly sampled between $30$ and $45$ degrees.
\subsubsection{Implementation Details}
\modelname is implemented using PyTorch on a single NVIDIA RTX 3090 GPU. The rendering resolution is set to 128$\times$128 in the first phase, and 512$\times$512 in both the second and third phases. We utilize the Adam optimizer with learning rates of 0.01 and 0.001 in the first and third phases, respectively, and the Adan optimizer with a learning rate of 0.01 in the second phase.
\subsubsection{3D Representation}
In the first optimization phase of \modelname, to enhance training and rendering efficiency, we employ multi-resolution hash grids and Instant NGP~\cite{muller2022instant}, parameterizing the scene's density and color through MLPs. In the second and third optimization phases of \modelname, to enhance the geometric and texture details of the low-resolution \nerf obtained in the first phase, we convert the neural field into Signed Distance Field (SDF) by setting an isosurface threshold of $10.0$ and adopt the high-resolution \dmtet~\cite{shen2021deep} as the 3D representation for subsequent geometric and texture refinement. \dmtet is a hybrid scene representation method that decouples the geometry and texture of 3D objects, explicitly modeling surfaces and synthesizing views through surface rendering, thus addressing the poor surface recovery in \nerf caused by coupling surface geometry learning with pixel color learning.
\subsubsection{Evaluation Metrics}
To facilitate evaluation, we render each 3D object from 120 viewpoints with uniformly distributed azimuth angles. We evaluate any-to-3D generation across three dimensions. For text-to-3D, we measure the CLIP R-Precision (CLIP-R) of images rendered from each 3D object using 136 text prompts from the DreamFusion gallery~\cite{poole2022dreamfusion} and compute the average. For image-to-3D, we calculate the CLIP similarity (CLIP-I) between the rendered images and the reference image and take the average. For audio-to-3D, due to the lack of suitable evaluation metrics, we propose the Audio-Rendering Cosine Consistency (ARCC) metric:
\begin{equation}
    \mathrm{ARCC} (A,I)=\mathrm{cos} (C(A),C(I) ),
\end{equation}
where $A$ and $I$ represent the audio prompts and rendered images, respectively, and $C$ denotes the \encodername.
This metric calculates the cosine similarity between the aligned embeddings of the audio and the rendered images using the \encodername. All three evaluation metrics measure the coherence and consistency between the prompt modalities and the generated 3D objects. Therefore, higher metric values indicate better alignment and similarity between the generated 3D objects and the modality prompts.

\subsection{Any-to-3D Generation}

In \figref{examples}, we present the generated results of \modelname. It can be observed that, when text, images, and audio are used as input prompts respectively, \modelname effectively processes the semantic information of each modality and integrates it into the 3D object generation and optimization process. As a result, the generated 3D objects align well with the modality prompts. Notably, for image input, our work differs from previous image-to-3D works in that \modelname focuses on \textbf{generation} rather than \textbf{reconstruction}, aiming to generate 3D objects that resemble the style of the input images, rather than producing a 1:1 reconstruction. For example, in the second row, second column of \figref{examples}, the input image is a frontal view of the Potala Palace, and the result generated by \modelname is a temple on a snowy mountain. This demonstrates that \modelname can effectively understand the layout and structure of objects in the image and generate imaginative 3D objects with a similar style. The results in \figref{examples} clearly show that, regardless of the type of modality prompt, \modelname is capable of generating high-fidelity 3D objects with detailed geometry and texture, while maintaining excellent 3D consistency. This demonstrates the effectiveness of hybrid diffusion supervision and the three-phase optimization approach.


\subsection{Comparison with Baselines}
As a pioneering work in Any-to-3D generation, we developed three baseline models for comparative analysis. These baselines use different diffusion models and loss functions for three-phase 3D object optimization. Firstly, ZeroStableCDS-3D employs Zero-1-to-3 and Stable unCLIP as 3D and 2D diffusion priors, respectively, with CDS loss for supervision across all phases. Secondly, StableSDS-3D uses only Stable unCLIP as the 2D diffusion prior and SDS loss for all phases. Lastly, ZeroSDS-3D uses only Zero-1-to-3 as the 3D diffusion prior and SDS loss for all phases.

\figref{comparison} presents a qualitative comparison. For the ZeroStableCDS-3D baseline, severe noise and even mode collapse are observed in the generated 3D objects, regardless of the modality used as the prompt condition, as shown in the “ZeroStableCDS-3D" column in \figref{comparison}. We hypothesize that this is due to the CDS loss being unsuitable for the initial optimization stage of 3D objects under different modality conditions. For the StableSDS-3D baseline, the generated 3D objects exhibit significant inconsistency and lack texture and geometric details, as shown in the “StableSDS-3D" column in \figref{comparison}. As for the ZeroSDS-3D baseline, the 3D objects generated have relatively accurate geometry and texture only from specific reference viewpoints. However, in non-reference views, this method struggles to “imagine" the detailed texture and geometry, showing a lack of generalization, as shown in the “ZeroSDS-3D" column in \figref{comparison}. 
In contrast, \modelname benefits from our hybrid diffusion supervision and the use of different loss functions across the three phases, allowing it to generate high-fidelity 3D objects with rich geometric and texture details, maintaining consistency and outperforming the other baselines.

The quantitative comparison results in \tabref{tab-comparison} demonstrate that our method surpasses the baselines in all three evaluation metrics—CLIP-R, CLIP-I, and ARCC. This further confirms that \modelname produces high-quality 3D objects that align well with the modality prompts, accurately capturing the semantic information, and highlights the superior performance of \modelname.

\begin{table}[!t]
    \centering
    \caption{Quantitative comparison of different models using CLIP-R, CLIP-I, and ARCC metrics.}
    \label{tab-comparison}
    \begin{tabular}{l|ccc}
        \toprule
        Model & CLIP-R & CLIP-I & ARCC \\
        \midrule
        ZeroStableCDS-3D & 0.4317 & 0.6723 & 0.3218 \\
        StableSDS-3D & 0.5583 & 0.8176 & 0.4044 \\
        ZeroSDS-3D & 0.4483 & 0.7267 & 0.3434 \\
        \midrule
        \modelname & \textbf{0.8050} & \textbf{0.8554} & \textbf{0.4860} \\
        \bottomrule
    \end{tabular}
\end{table}

\subsection{Comparison with the SOTA Text-to-3D Methods}
\begin{figure*}[p]
\centering
\includegraphics[width=0.9\textwidth]{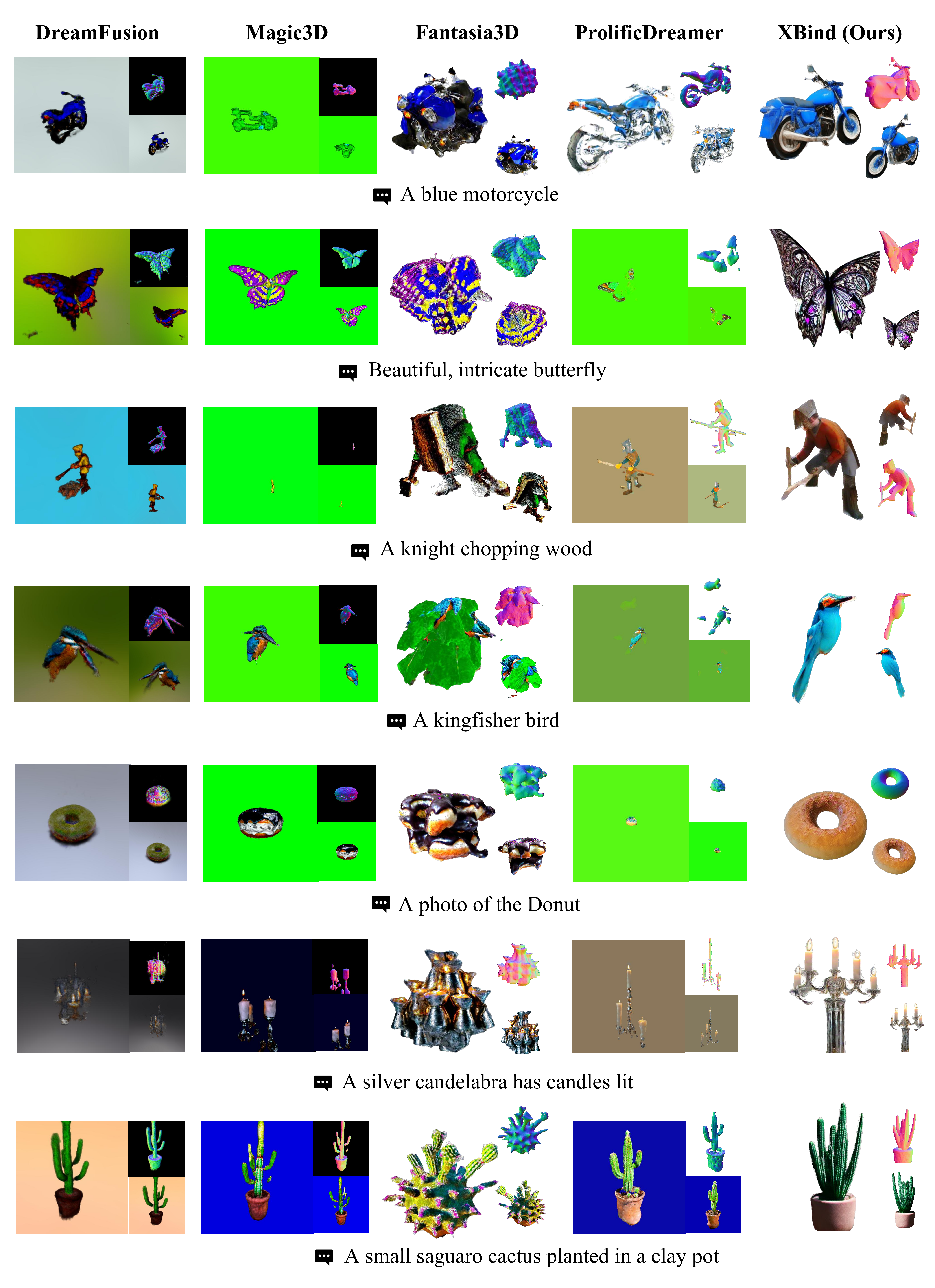} 
\caption{Qualitative comparison with SOTA methods in the text-to-3D domain.}
\label{text-to-3d-comparison}
\end{figure*}

For the text-to-3D task, we used the same text prompts to qualitatively and quantitatively compare the results of \modelname with state-of-the-art (SOTA) methods. In this comparison, we included four different advanced text-to-3D methods: DreamFusion~\cite{poole2022dreamfusion}, Magic3D~\cite{lin2023magic3d}, Fantasia3D~\cite{chen2023fantasia3d}, and ProlificDreamer~\cite{wang2024prolificdreamer}. To ensure fairness, the results of these four SOTA methods were obtained using implementations from the open-source repository threestudio~\cite{threestudio2023}. The comparison results are shown in \figref{text-to-3d-comparison} and \tabref{tab-text-comparison}. 
In \figref{text-to-3d-comparison}, it can be observed that the 3D objects generated by \modelname exhibit higher quality, richer geometric and texture details, and better 3D consistency compared to other SOTA methods. As shown in \tabref{tab-text-comparison}, \modelname achieved higher CLIP-R scores, indicating that it captures the semantic information from the text prompts more accurately and generates 3D objects that align better with the textual descriptions. Overall, \modelname outperforms the four SOTA methods in terms of geometric structure, texture quality, and semantic capture, highlighting the effectiveness of each module in \modelname and demonstrating its ability to generate high-fidelity textured meshes.

\begin{figure*}[t]
\centering
\includegraphics[width=1.0\textwidth]{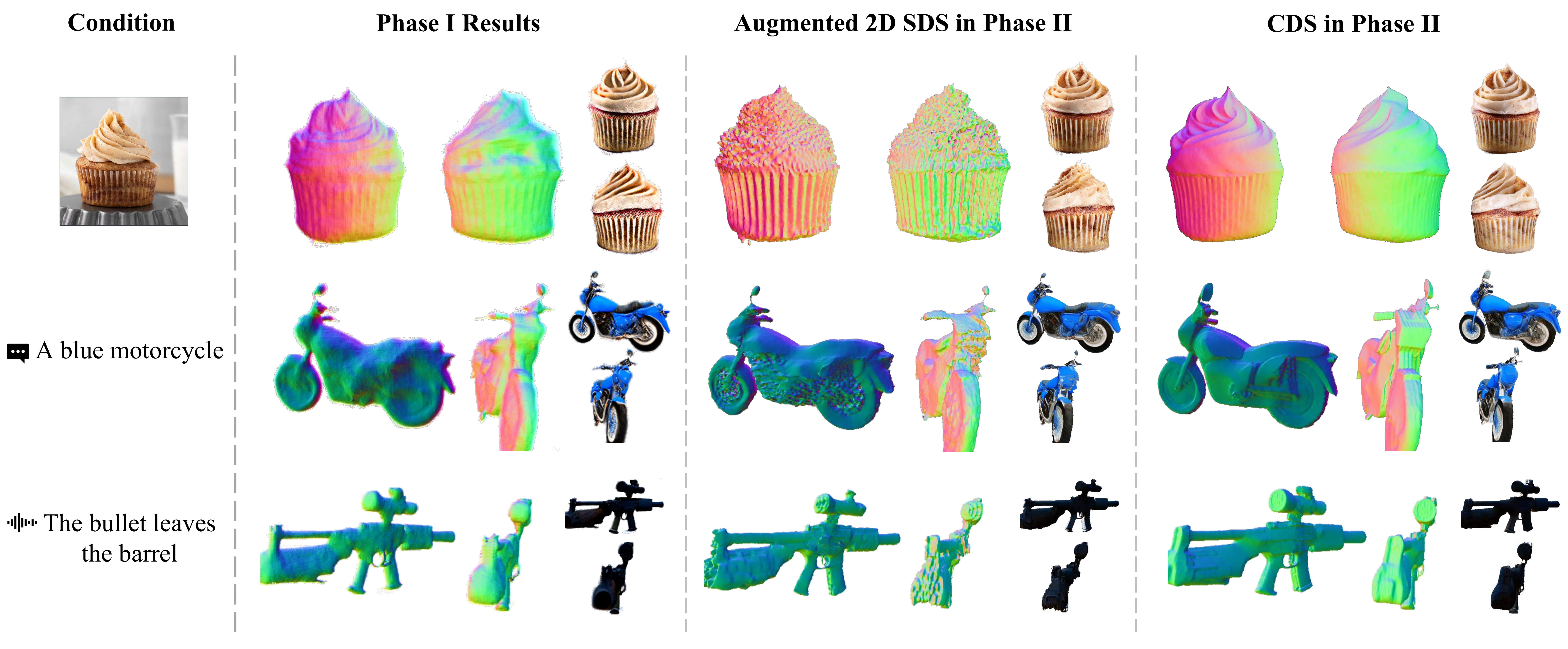} 
\caption{Results of using different 2D diffusion model supervision losses in \textit{Phase II}.}
\label{sup_phase_II}
\end{figure*}

\begin{figure*}[t]
\centering
\includegraphics[width=1.0\textwidth]{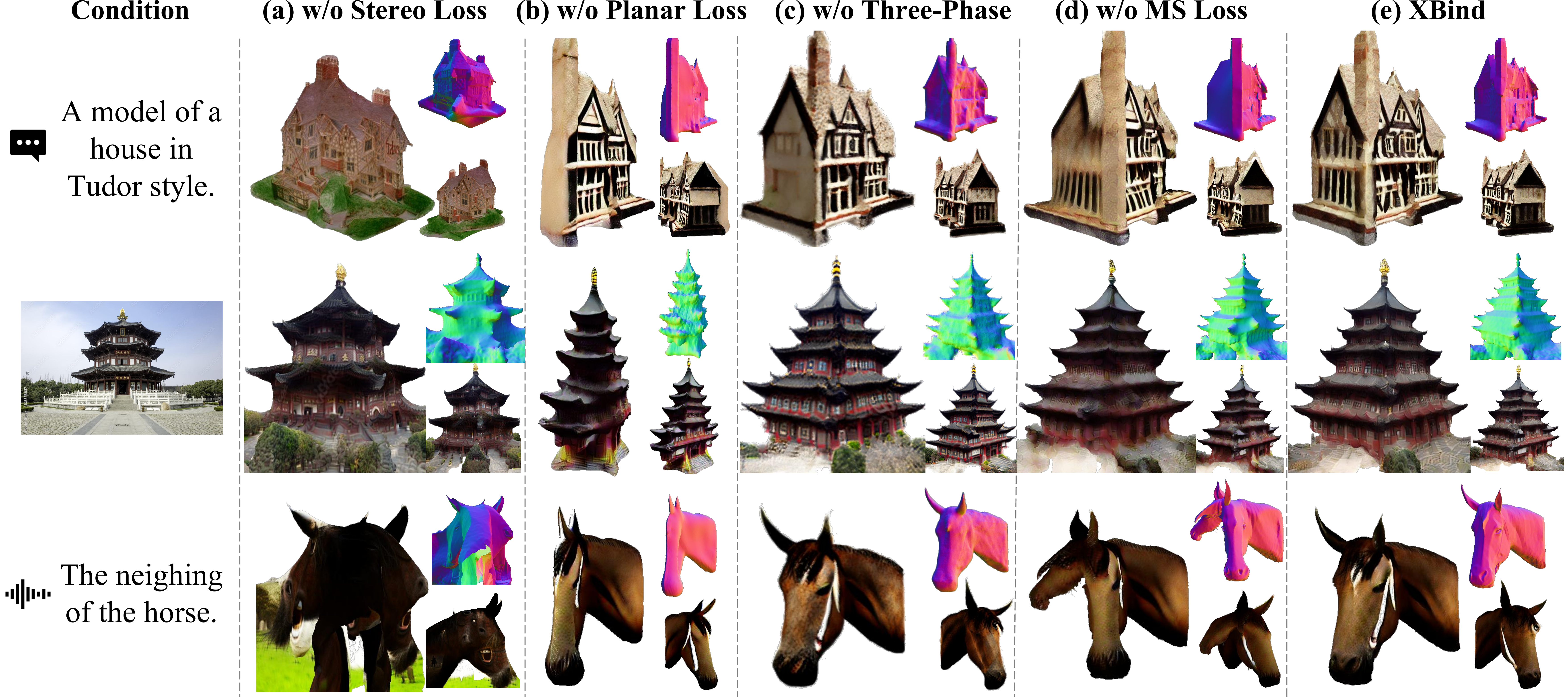} 
\caption{Ablation study of \modelname. The first row represents text-to-3D, the second row represents image-to-3D, and the third row represents audio-to-3D.}
\label{ablation}
\end{figure*}

\subsection{In-Depth Discussion of the Supervision Losses Used in \textit{Phase II}}
Our model, \modelname, utilizes a three-phase optimization framework for generating high-quality 3D objects. In the second phase of our model's optimization, \emph{i.e.} the geometry refinement phase, we employed the CDS loss as the supervision loss for the 2D diffusion model, which differs from the augmented 2D SDS loss used in the first and third phases. The reason for this change is that we found the augmented 2D SDS loss was unable to produce correct geometric details during this phase. As shown in the “Augmented 2D SDS in Phase II" column of~\figref{sup_phase_II}, the generated 3D objects exhibited many undesirable “bumps" on their surfaces, resulting in poor overall quality and failing to achieve the goal of geometric refinement. Conversely, when using CDS as the supervision loss, the model successfully handled the geometry of the 3D objects, as illustrated in the “CDS in Phase II" column of ~\figref{sup_phase_II}. Compared to the results from the first phase of optimization, shown in the “Phase I Results" column of ~\figref{sup_phase_II}, the use of CDS loss facilitated the refinement of geometric details. Therefore, in this phase, we combined CDS loss with \nameloss to provide pixel-level planar supervision, as shown in (\ref{planar-cds}), and further incorporated spatial-level stereoscopic supervision to refine the geometry of the 3D objects.
\begin{equation}
    \mathcal{L}_{planar}=\omega_{ms}\mathcal{L}_{MS}+ \omega_{cds}\mathcal{L}_{CDS}
    \label{planar-cds}
\end{equation}
The results shown in the “Augmented 2D SDS in Phase II" and “CDS in Phase II" columns of~\figref{sup_phase_II} were both initialized with the same results from the first phase (“Phase I Results" column), and all configurations were kept the same except for the supervision loss (CDS loss or augmented 2D SDS loss) of the 2D diffusion model. In the “Augmented 2D SDS in Phase II" and “CDS in Phase II" columns of \figref{sup_phase_II}, the rightmost colored image of each generated object represents the final result after the \textit{Phase III} texture refinement, based on the geometrically refined 3D objects generated in the \textit{Phase II} of that column.

\begin{table}[]
    \centering
    \caption{Quantitative comparison with SOTA methods in text-to-3D generation.}
    \label{tab-text-comparison}
    \begin{tabular}{l|c}
        \toprule
        Model & CLIP-R \\
        \midrule
        DreamFusion~\cite{poole2022dreamfusion} & 0.6576\\
        Magic3D~\cite{lin2023magic3d} & 0.6181 \\
        Fantasia3D~\cite{chen2023fantasia3d} & 0.5250 \\
        ProlificDreamer~\cite{wang2024prolificdreamer} & 0.7319 \\
        \midrule
        \modelname & \textbf{0.8167} \\
        \bottomrule
    \end{tabular}
\end{table}

\subsection{Ablation Study}
We conducted an ablation study on the various modules in \modelname to evaluate their effectiveness. The results are shown in \figref{ablation}. 

(a) Without spatial-level stereoscopic supervision, the model lacks spatial priors, resulting in incorrect geometry and textures, and significant 3D inconsistencies in the generated 3D objects. As shown in the first generated object in the “(a) w/o Stereo Loss" column of \figref{ablation}, the model fails to generate a regular house shape from the text prompt “A model of a house in Tudor style.” Instead, it produces an object with multiple surrounding walls and unrealistic textures. This highlights the crucial role of spatial-level stereoscopic supervision in ensuring the geometric accuracy and consistency of both geometry and textures in the generated 3D models.

(b) Without pixel-level planar supervision, the model struggles to generalize, leading to texture loss and incorrect geometry in non-reference viewpoints. As seen in the second generated object in the “(b) w/o Planar Loss” column of \figref{ablation}, the side of the tower (a non-reference view) exhibits incorrect texture generation, lacking expected details such as “windows,” and the overall geometry of the tower becomes flattened, deviating from a typical tower structure. This demonstrates that pixel-level planar supervision improves the generalization ability of 3D object generation, enabling high-quality results from different viewpoints.

(c) Relying solely on the first phase for 3D optimization results in 3D objects with a lack of geometric details and blurred textures, leading to a lack of realism. As shown in the first generated object in the “(c) w/o Three-Phase” column of \figref{ablation}, compared to the result generated by \modelname in the final column, the house generated by \modelname exhibits more accurate geometric details, such as the shape of the chimney. At the same time, the house generated in the first phase lacks well-defined side textures, whereas \modelname generates textures consistent with the overall style of the house. This illustrates that the three-phase optimization framework ensures the generated 3D objects have richer geometric and texture details, improving the quality and fidelity of the results.

(d) Without \nameloss, the model fails to incorporate prompt information effectively during 3D generation, resulting in inconsistent and less detailed 3D objects. As shown in the second and third generated objects in the “(d) w/o MS Loss” column of \figref{ablation}, the geometric structure of the tower features multiple eaves corners, accompanied by a loss of texture details, while the horse exhibits a multi-head phenomenon. These issues indicate that, without \nameloss, the model cannot integrate the prompt information well for 3D generation, highlighting the importance of \nameloss in ensuring 3D consistency and high-quality geometry and textures.

(e) As shown in the final column of \figref{ablation}, with all modules integrated, \modelname generates high-fidelity, high-quality textured meshes with rich geometric and texture details, while maintaining good 3D consistency. This success is attributed to the significant contributions of each module.


\section{Limitations}
Although \modelname can generate high-fidelity 3D objects using various modalities, the generated results rely on the priors of two diffusion models. Therefore, inherent limitations of the diffusion models, such as the inability to render clear text, poor performance on more complex tasks involving compositionality, and potential inaccuracies in generating faces and people, can affect the output of our model. Furthermore, similar to~\cite{chen2023fantasia3d}, our model focuses on generating 3D objects rather than complete 3D scenes.

\section{Conclusion}
In this work, we propose \modelname, a pioneering unified framework for any-to-3D generation. Current single-modality-to-3D generation models are limited to generating 3D objects based on a single modality. When dealing with different types of modality data, frequent switching between models or converting modalities before generation is required. This approach is not only time-consuming and labor-intensive but also prone to information loss during modality conversion. To address these issues, we integrate the \encodername with diffusion models to achieve any-to-3D generation. Specifically, we introduce a new loss function, \fullnamelossb, which aligns the modality prompt embeddings from the \encodername with the CLIP embeddings of 3D rendered images. \fullnamelossb effectively incorporates the semantic information from different modality prompts into the 3D generation process. Additionally, we propose a Hybrid Diffusion Supervision strategy, which combines pixel-level planar supervision and spatial-level stereoscopic supervision, leveraging the strengths of both 2D and 3D diffusion models to generate consistent and diverse 3D objects. To further enhance generation quality, we introduce a coarse-to-fine Three-Phase optimization framework. In each phase of this framework, different supervision methods from Hybrid Diffusion Supervision are applied to improve the quality and fidelity of 3D object generation. In \modelname, each module plays a crucial role, and by combining them, \modelname is capable of accepting any modality as input and generating high-quality 3D objects with detailed geometry and texture that correspond to the input modality description.


\bibliographystyle{IEEEtran}
\bibliography{IEEEabrv,IEEEexample}

\end{document}